\documentclass[twoside]{article}

\usepackage{graphicx}
\usepackage{algorithm}
\usepackage{algorithmic}
\usepackage{amsfonts, amsmath}
\usepackage{multirow}
\usepackage{natbib}
\usepackage{amssymb}
\usepackage[preprint]{aistats2026}
\begin{document}

%

%

\twocolumn[

\aistatstitle{FedOAP: Organ Agnostic Tumor Segmentation Using Personalized Federated Learning}

\aistatsauthor{Ishmam Tashdeed* \And Md. Atiqur Rahman* \And  Sabrina Islam \And Md. Azam Hossain}

\aistatsaddress{Islamic University of Technology \\ $^{\ast}$\textit{Equal Contribution}} ]

\begin{abstract}
Personalized federated learning (PFL) possesses the unique capability of preserving data confidentiality among clients while tackling the data heterogeneity problem of non-independent and identically distributed (Non-IID) data. Its advantages have led to widespread adoption in domains such as medical image segmentation. However, the existing approaches mostly overlook the potential benefits of leveraging shared features across clients, where each client contains segmentation data of different organs. In this work, we introduce a novel personalized federated approach for organ agnostic tumor segmentation (FedOAP), that utilizes cross-attention to model long-range dependencies among the shared features of different clients and a boundary-aware loss to improve segmentation consistency. FedOAP employs a decoupled cross-attention (DCA), which enables each client to retain local queries while attending to globally shared key-value pairs aggregated from all clients, thereby capturing long-range inter-organ feature dependencies. Additionally, we introduce perturbed boundary loss (PBL) which focuses on the inconsistencies of the predicted mask's boundary for each client, forcing the model to localize the margins more precisely. We evaluate FedOAP on diverse tumor segmentation tasks spanning different organs. Extensive experiments demonstrate that FedOAP consistently outperforms existing state-of-the-art federated and personalized segmentation methods.
\end{abstract}

\section{Introduction}
\begin{figure}[t]
    \centering
    \includegraphics[width=\columnwidth]{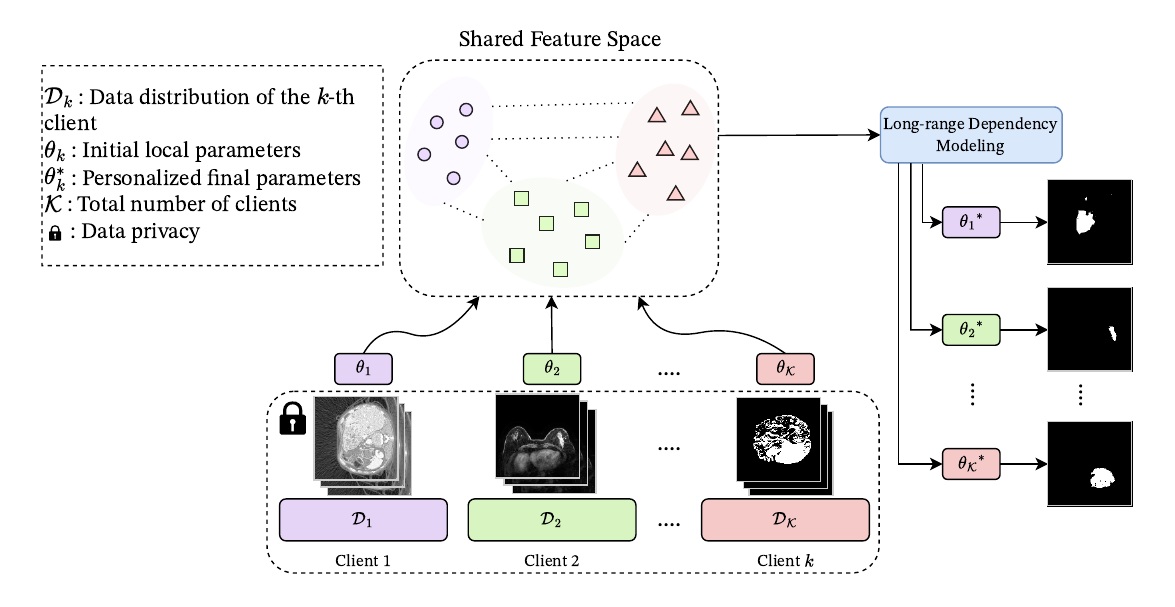} 
    \caption{Motivation for FedOAP. FedOAP aims to learn a shared feature representation across clients by aggregating informative tokens from different organs while preserving client-specific data privacy. This enables improved tumor segmentation by leveraging inter-organ knowledge and producing personalized predictions guided by cross-client feature dependencies.}
    \label{fig:inspo}
\end{figure}

Federated Learning (FL) is a decentralized machine learning paradigm that enables collective model training across multiple data sources without the need to share raw data \citep{li2020federated}. By ensuring that data remains local to its source, FL inherently adheres to privacy regulations and data protection standards \citep{voigt2017eu}. This makes FL particularly well-suited for sensitive domains such as healthcare \citep{horst2025federated, kaissis2020secure, kwak2023role}. However, non-independent and identically distributed (Non-IID) data or heterogeneous data poses as a great hurdle for most FL methods \citep{zhao2018federated}. A universally shared global model is usually insufficient to tackle the diverse types of data encountered across different clients. Heavily heterogeneous data may introduce client drift which severely degrades the model's performance \citep{karimireddy2020scaffold}. Specially in the domain of medical image analysis, it is common to have inter-client inconsistencies as individual clients often address distinct clinical tasks, patient populations, cultural contexts, and imaging equipment \citep{guan2024federated}.

Personalized federated learning (PFL) tries to address this problem by maintaining distinct local models for each client that conforms to its unique distribution while keeping the data localized \citep{tan2022towards}. This is achieved by a variety of methods such as regularization of local and global weight divergence \citep{karimireddy2020scaffold, li2020fedprox}, meta-learning \citep{fallah2020personalized, t2020personalized}, transfer-learning \citep{afzali2025phifl, tan2023transfer}, decoupling parameters into shared and private segments \citep{arivazhagan2019federated}, knowledge distillation \citep{li2019fedmd, lin2020ensemble}, clustering clients based on similar tasks or data distributions \citep{briggs2020federated, sattler2020clustered}, among others. In medical image segmentation, a field often bound by strict data privacy policies, PFL is heavily relied upon. The most pervasive strategy is to share the feature extractor while training a separate mask prediction head for each client to accommodate local data distributions \citep{wang2023feddp, xie2024pflfe, wang2022personalizing, jiang2023iop, liu2025pcrfed}.

Despite recent strides made in tumor segmentation using PFL, existing approaches have largely overlooked the potential benefits of leveraging cross-organ information to enhance segmentation performance. In a federated setting, clients often possess segmentation data regarding anatomically different organs such as brain, liver, or breast. We hypothesize that a shared feature extractor with appropriate personalization can greatly benefit the segmentation process across all the clients. However, most existing methods are not well suited to accommodate data originating from different organs and often from different modalities \citep{madni2023federated, babar2024investigating}. This can potentially be handled by modeling long-range dependencies via cross-attention mechanisms, enabling the aggregation of global contextual information across the feature space of the clients. Furthermore, introducing personalization during feature extraction across clients not only aids in the adaption of the local models to the heterogeneous data distribution but also preserves data privacy \citep{wang2023feddp}. For effective client personalization local fine-tuning is typically required, where each client further trains its model for a few epochs on its private dataset. During this step, inconsistencies in the predicted masks greatly degrades the segmentation performance \citep{wang2023feddp, wang2022personalizing}. Moreover, the few methods that attempt to resolve this often introduce substantial computational or memory overhead.

In this work, we propose FedOAP, a novel personalized federated learning framework for organ-agnostic tumor segmentation, which integrates cross-organ aggregated feature extraction with a novel perturbed boundary loss to enhance both generalization and boundary precision across heterogeneous clients. To handle distinct heterogeneous organ data from different clients, FedOAP utilizes a shared encoder-decoder while introducing decoupled cross-attention (DCA) which decouples query-key embedding layers and employs cross-attention on aggregated key-value pairs from all clients. Inspired by \citep{wang2023feddp}, keeping the query layer local to each client, we ensure data privacy as query embeddings generally represent the features of the local data itself. Aggregating key-value embedding pairs across multiple clients enables the local query to attend to supportive features from different organs. Additionally, FedOAP adheres to established practices in PFL by incorporating a client-specific spatial adapter, while introducing a novel perturbed boundary loss (PBL) to enhance segmentation accuracy along proposed tumor boundaries in the fine-tuning process. PBL quantifies pixel-level inconsistencies between local predictions and ground truth and injects stochastic noise into the erroneous boundary regions. By formulating a loss based on these perturbed discrepancies, we direct greater supervision focus toward critical boundary pixels, thereby promoting the learning of more robust and precise boundary representations.

In general, our contributions are as follows:
\begin{itemize}
    \item We propose a novel PFL framework for tumor segmentation specifically designed to handle segmentation tasks across heterogeneous organ datasets named FedOAP.
    
    \item We introduce DCA, a mechanism that decouples the query and key-value embedding layers. The query remains local to each client, while the aggregated key-value pairs from all clients are used to enable cross-organ feature fusion through cross-attention.

    \item In addition, we propose PBL to improve proposed tumor mask accuracy. It identifies inconsistencies between predicted and ground truth boundaries while injecting stochastic noise into these regions, and formulates a targeted loss to enhance supervision on boundary pixels, thereby improving segmentation robustness and precision during client side fine-tuning.
\end{itemize}

We conduct comprehensive experiments on three tumor segmentation tasks from different organs: brain, liver, and breast. FedOAP achieves superior performance compared to existing state-of-the-art PFL methods, demonstrating its strong generalization capabilities and its effectiveness in handling inter-organ variability. We provide related work in Section \ref{sec:related_work} of the supplementary document.

\begin{figure*}[t]
    \centering
    \includegraphics[width=\textwidth]{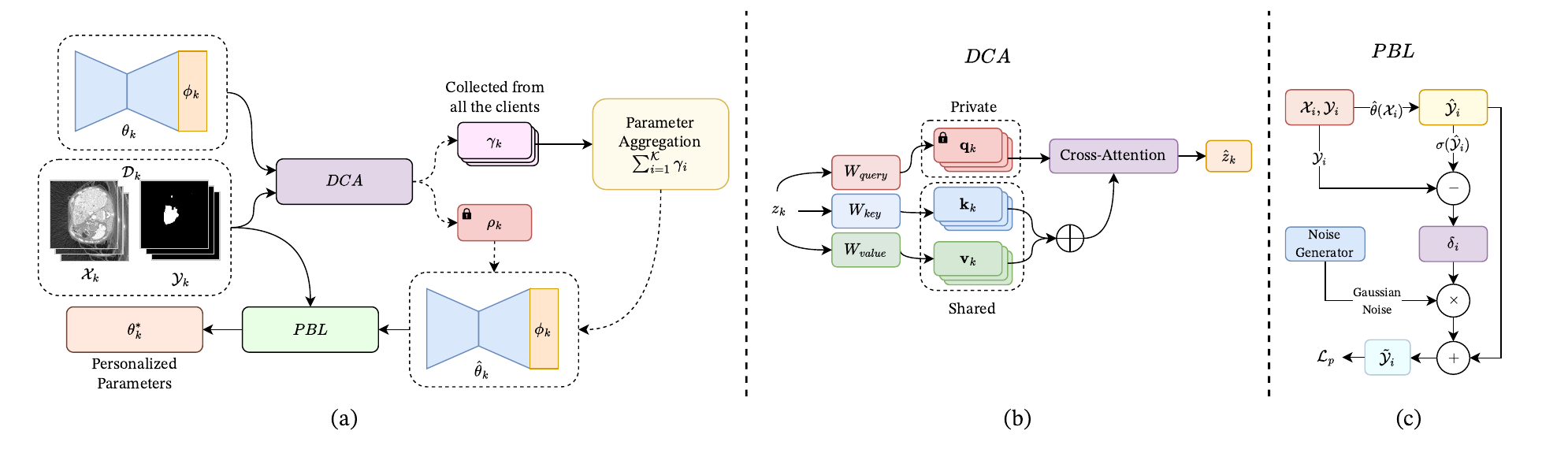} 
    \caption{(a) Overview of the proposed FedOAP framework. Each client $k \in \mathcal{K}$ trains a local model $\theta_k$ (composed of an encoder-decoder architecture with a client-specific spatial adapter $\phi_k$) on its private dataset $\mathcal{D}_k$ for $\mathcal{T}$ rounds. During this phase, DCA separates the shared parameters $\gamma_k$ for global aggregation. Finally, each client fine-tunes its model using PBL, resulting in the personalized parameters ${\theta}_k^{\ast}$. (b) The proposed Decoupled Cross-Attention (DCA) mechanism. The private query embeddings $\mathbf{q}_k$ attend to the concatenated key-value $(\mathbf{k}_k, \mathbf{v}_k)$ pairs, capturing cross-organ, long-range dependencies without exposing private client representations. (c) The Perturbed Boundary Loss (PBL) mechanism. The inconsistencies between predicted masks $\hat{\mathcal{Y}}_i$ and ground truth $\mathcal{Y}_i$ is used as a guide to inject noise, reinforcing supervision on uncertain regions to improve boundary precision.}
    \label{fig:framework}
\end{figure*}

\section{Methodology} \label{sec:methodology}
    \subsection{Problem Definition}
    Image segmentation is defined as a dense pixel-wise classification problem where given an input image, the model classifies each pixel to a predefined semantic class label. In the context of our binary tumor segmentation task, each pixel is classified as either belonging to a tumor region or the background. The trained segmentation model $f_{\theta}(\mathcal{X}) \rightarrow \mathcal{Y}$, parameterized by $\theta$, takes an image $x_i \in \mathbb{R}^{H \times W \times C}$ and outputs a segmentation mask $y_i \in \{0, 1\}^{H \times W}$. Here, $H$, $W$, and $C$ are the height, width, and number of channels respectively. Thus, the segmentation task can be formulated as:
    \begin{equation}
        \label{eq:SegDefinition}
        f_{\theta}(x_i) = \underset{y_i \in \mathcal{Y}}{\operatorname{argmax}} \mathbb{P}(y_i|x_i; \theta)
    \end{equation}

    We consider a federated setting with $\mathcal{K}$ clients, each possessing a data distribution $\mathcal{D}_{k \in \mathcal{K}}$ corresponding to a distinct organ. For the $k$-th client, a set of $N_k$ image-mask pairs, denoted as $\{x_{i, k}, y_{i, k}\}_{i=1}^{N_k}$ constitute the non-iid dataset such that $\{x_{i, k}, y_{i, k}\} \sim \mathcal{D}_k$. In classical FL scenarios such as \citep{mcmahan2017communication}, the goal is to learn a shared model $\theta$ in $\mathcal{T}$ rounds by initially training on a client's local distribution $\mathcal{D}_k$ and aggregating them each round as:
    \begin{equation}
        \label{eq:FedAVG}
        {\theta}^{t+1} = {\theta}^t - \alpha \mathbb{E}_{k} \mathbb{E}_{(x_{i, k}, y_{i, k}) \sim \mathcal{D}_k} \nabla \mathcal{L}_k(f_{\theta}(x_{i, k}), y_{i, k})
    \end{equation}
    Here, $\alpha$ is the learning rate, $t \in \{0, 1, \dots, \mathcal{T}-1\}$ denotes the communication round, and $\mathcal{L}_k$ is the objective function for the $k$-th client. However, a single shared model $\theta$ yields suboptimal performance due to highly heterogeneous data distribution across clients. PFL tackles this problem by learning $k$ separate models denoted as $\{{\theta}_k\}_{k=1}^{\mathcal{K}}$, where each ${\theta}_k$ aims to fit the client's distribution $\mathcal{D}_k$.

    \subsection{Framework Overview}
    Inspired by insights discussed in \citep{bengio2013representation}, we aim to utilize a cross-organ shared feature representation to enhance the performance of tumor segmentation in heterogeneous federated settings. Our proposed framework, as shown in Figure \ref{fig:framework}, is composed of two sequential stages: \emph{federated representation alignment} and \emph{client-specific personalization}, where the proposed decoupled cross-attention (DCA) and perturbed boundary loss (PBL) are performed respectively. During federated representation alignment, each client $k$ maintains a local model ${\theta}_k$, which is trained on its private dataset $\mathcal{D}_k$ over $\mathcal{T}$ communication rounds. The model parameters are partitioned into two disjoint subsets: shared parameters ${\gamma}_k$ and personalized parameters ${\rho}_k$ such that ${\theta}_k = {\gamma}_k \cup {\rho}_k$. At each round $t$, the shared parameters are aggregated through averaging: ${\gamma}^{t+1}=\frac{1}{\mathcal{K}}\sum_{k=1}^{\mathcal{K}}{\gamma}_{k}^{t}$ following \citep{mcmahan2017communication}. These aggregated parameters are broadcast back to the clients and merged with the personalized parameters to create a set of preliminary models $\hat{\theta}_k^{t+1}$ such that: $\hat{\theta}_k^{t+1} = {\gamma}^{t+1} \cup {\rho}_k^t$. The client-specific personalization phase, fine-tunes each local model $\hat{\theta}_k$ on its dataset without further parameter sharing by minimizing a composite loss. This process yields the final personalized set of models $\{{\theta}_k^{\ast}\}_{k=1}^{\mathcal{K}}$, each tailored to its local data distribution while benefiting from cross-client feature alignment.

    \begin{algorithm}[t]
    \caption{FedOAP Algorithm.}
    \label{alg:FedOAP}
    \begin{algorithmic}[1] 
        \REQUIRE Local datasets $\{\mathcal{X}_k, \mathcal{Y}_k\}_{k=1}^{\mathcal{K}} \sim \mathcal{D}_k$, 
        total number of rounds $\mathcal{T}$, number of clients $\mathcal{K}$, 
        number of fine-tuning epochs $N_\epsilon$.
        \ENSURE Local parameters ${\theta}_{k}^{\ast}$.
        \STATE Initialize personalized parameters for each client ${\rho}_k^{t=0}$
        \STATE Initialize shared parameters for all clients ${\gamma}^{t=0}$
        \FOR{$t=0, 1, \dots, \mathcal{T}-1$}
            \FOR{$k=1, 2, \dots, \mathcal{K}$}
                \STATE ${\gamma}_k^{t}, {\rho}_k^{t+1} \leftarrow \operatorname{\mathbf{DCA}}(\{\mathcal{X}_k, \mathcal{Y}_k\}, {\gamma}^t, {\rho}_k^t)$
            \ENDFOR
            \STATE ${\gamma}^{t+1} \leftarrow \frac{1}{\mathcal{K}} \sum_{i=1}^{\mathcal{K}}{\gamma}_{i}^{t}$
        \ENDFOR
        \FOR{$k=1, 2, \dots, \mathcal{K}$}
            \STATE $\hat{\theta}_k \leftarrow {\gamma}^{\mathcal{T}} \cup {\rho}_k^{\mathcal{T}}$
        \ENDFOR
        \FOR{$k=1, 2, \dots, \mathcal{K}$}
            \FOR{$i=0, 1, \dots, N_{\epsilon}$}
                \STATE ${\theta}_k^{\ast} \leftarrow \operatorname{\mathbf{PBL}}(\{\mathcal{X}_k, \mathcal{Y}_k\}, \hat{\theta}_k)$
            \ENDFOR
        \ENDFOR
        \STATE \textbf{return} $\{{\theta}_k^{\ast}\}_{k=1}^{\mathcal{K}}$
    \end{algorithmic}
    \end{algorithm}

    \subsection{Federated Representation Alignment}
    To effectively capture long-range dependencies across heterogeneous client representations, we slightly modify the self-attention mechanism originally introduced in \citep{vaswani2017attention}. Prior work, such as \citep{wang2023feddp}, has demonstrated the efficacy of self-attention in modeling long-range feature interactions while simultaneously accommodating inter-client variability in federated settings. Concurrently, methods including \citep{li2021fedbn, arivazhagan2019federated, dai2024federated} have emphasized the importance of personalizing specific model components to better adapt to client-specific data distributions. Let $z \in \mathbb{R}^{n \times d}$ denote the input to a self-attention block. Here, $n$ is the sequence length and $d$ is the feature embedding dimension. The \emph{query} ($\mathbf{q} \in \mathbb{R}^{n \times d_q}$), \emph{key} ($\mathbf{k} \in \mathbb{R}^{n \times d_k}$), and \emph{value} ($\mathbf{v} \in \mathbb{R}^{n \times d_v}$) are calculated as: $\mathbf{q}=zW_{query}, \mathbf{k}=zW_{key}, \mathbf{v}=zW_{value}$, where $W_{query}, W_{key}, W_{value} \in \mathbb{R}^{d \times d}$ are learnable projection matrices. For simplicity, we assume $d_q = d_k = d_v = d$. The scaled dot-product attention $A(z)$ is calculated as:
    \begin{equation}
        \label{eq:selfAttn}
        A(\mathbf{q}, \mathbf{k}, \mathbf{v}) = \operatorname{softmax} \left(\frac{\mathbf{q}\mathbf{k}^T}{\sqrt{d}}\right)\mathbf{v}
    \end{equation}

    Following \citep{wang2023feddp}, we propose \emph{Decoupled Cross-Attention} (DCA). In this formulation, each client $k$ maintains a personalized query embedding $\mathbf{q}_k$, while attending to globally shared key-value embeddings $(\mathbf{k}_{shared}, \mathbf{v}_{shared})$ concatenated from all clients as $\mathbf{k}_{shared} = concat(\mathbf{k}_1, \mathbf{k}_2, \dots, \mathbf{k}_{\mathcal{K}}) \in \mathbb{R}^{N \times d}$ and $\mathbf{v}_{shared} = concat(\mathbf{v}_1, \mathbf{v}_2, \dots, \mathbf{v}_{\mathcal{K}}) \in \mathbb{R}^{N \times d}$, where $N$ is the total number of aggregated tokens. Additionally, to help our models adapt to the local distributions, we append a personalized spatial adapter ${\phi}_{k \in \mathcal{K}}$ to the decoder of each client. Following established notation, the personalized parameters for client $k$ are ${\rho}_k = \{\mathbf{q}_k, {\phi}_k\}$, while the shared parameters are ${\gamma}_k$, such that ${\gamma}_k \cap {\rho}_k = \emptyset$. During training, client-specific parameters ${\rho}_k$ are updated as:
    \begin{equation}
        \label{eq:FedAVGLocal}
        {\rho}_k^{t+1} = {\rho}_k^t - \alpha \mathbb{E}_{(x_{i, k}, y_{i, k}) \sim \mathcal{D}_k} \nabla \mathcal{L}_k(f_{{\theta}_k}(x_{i, k}), y_{i, k})
    \end{equation}
    In parallel, the shared parameters ${\gamma}_k$ are collaboratively optimized across all clients as:
    \begin{equation}
        \label{eq:FedAVGGlobal}
        {\gamma}_k^{t+1} = {\gamma}_k^t - \alpha \mathbb{E}_{k} \mathbb{E}_{(x_{i, k}, y_{i, k}) \sim \mathcal{D}_k} \nabla \mathcal{L}_k(f_{{\theta}_k}(x_{i, k}), y_{i, k})
    \end{equation}
    After completing $\mathcal{T}$ communication rounds, each client obtains an intermediate parameters $\hat{\theta}_k$, which are further updated in the subsequent phase.

    \subsection{Perturbation Guided Calibration} \label{sec:loss}
    Different clients possess varying modalities and sources of data and the absence of a unified imaging protocol being enforced across all clients introduces domain shift \citep{zhu2024adaptive, pan2025adaptive}. This introduces minor inconsistencies between the predicted mask and the ground truth which greatly affects a model's performance. Degradation caused by this is often more noticeable when the clients contain remarkably heterogeneous data such as from different organs. Additionally, inconsistencies can often guide the model towards improving segmentation results \citep{ji2021learning, wang2023feddp, wang2022personalizing}. Focusing on these discrepancies during the local fine-tuning phase, our models can greatly acclimate to the client's unique distribution. Emboldened by this, we propose a \emph{Perturbed Boundary Loss} (PBL), which aims to adjust the model's prediction guided by the differences between the predicted and the actual mask in a robust manner.

    Given intermediate parameters $\hat{\theta}_k$, the model infers each sample $(x_{i, k}, y_{i, k}) \sim \mathcal{D}_k$ to obtain predicted masks $\hat{y}_{i, k}$. The inconsistency mask $\delta_{i, k} \in \{0, 1\}^{H \times W}$ is computed using:
    \begin{equation}
        \label{eq:predictedDiff}
        \delta_{i, k} = 
        \begin{cases}
            1 & \text{if } \left|y_{i, k} - \sigma(\hat{y}_{i, k})\right| > \tau \\
            0 & \text{otherwise}
        \end{cases}
    \end{equation}
    Here, we pass $\hat{y}_{i, k}$ through the sigmoid function $\sigma$ and $\tau \in (0, 1)$ is a cut-off threshold value. We introduce additive noise $\epsilon_{i, k} \sim \mathcal{N}$ to the regions indicated as inconsistencies from the inconsistency mask $\delta_{i, k}$ to get the noisy mask: $\tilde{y}_{i, k} = \hat{y}_{i, k} + \delta_{i, k} \cdot \epsilon_{i, k}$. Here, $\mathcal{N}$ is the gaussian distribution with mean of $0$ and variance of $0.1$. We simultaneously compute two losses: Segmentation loss $\mathcal{L}_s$ using the clean predicted masks $\{\hat{y}_{i, k}\}_{k=1}^{\mathcal{K}}$ and Perturbed loss $\mathcal{L}_p$ with the noisy masks $\{\tilde{y}_{i, k}\}_{k=1}^{\mathcal{K}}$. The composite objective function is defined as $\mathcal{L} = (1-\lambda) \mathcal{L}_s + \lambda \mathcal{L}_p$, where $\lambda \in [0, 1]$ is the trade-off parameter between the two objectives. The complete algorithm regarding the FedOAP framework is detailed in Algorithm \ref{alg:FedOAP}.

\begin{table*}[t]
    \caption{Comparison of FedOAP with state-of-the-art methods across multiple datasets. Reported values denote the mean and standard deviation over five runs for each client, along with the average across clients. The best performing results are highlighted in bold.}
    \label{tab:results}
    \begin{center}
    \begin{tabular}{c|ccccccc}
    \hline
    \multirow{2}{*}{Method} & \multicolumn{2}{c}{BreastDM} & \multicolumn{2}{c}{BraTS} & \multicolumn{2}{c}{LiTS} & Average(\%) $\uparrow$ \\ \cline{2-8} 
                            & Dice(\%) $\uparrow$        & Variance       & Dice(\%) $\uparrow$       & Variance     & Dice(\%) $\uparrow$      & Variance     &         \\ \hline
    FedAvg                  & $50.91$     & $\pm3.62$      & $1.28$     & $\pm1.17$    & $9.71$    & $\pm0.95$    & $20.63$ \\
    FedAvgM                 & $43.29$     & $\pm4.66$      & $0.96$     & $\pm1.48$    & $10.66$   & $\pm1.1$     & $18.30$ \\
    FedAdagrad              & $4.51$      & $\pm0.27$      & $35.16$    & $\pm6.01$    & $3.63$    & $\pm0.19$    & $14.43$ \\ \hline
    FedPer                  & $59.91$     & $\pm23.47$     & $19.15$    & $\pm23.28$   & $42.06$   & $\pm24.12$   & $40.37$ \\
    FedRep                  & $85.92$     & $\pm1.99$      & $75.56$    & $\pm19.09$   & $49.75$   & $\pm2.94$    & $70.41$ \\
    FedDP                   & $91.47$     & $\pm0.72$      & $94.04$    & $\pm0.56$    & $83.88$   & $\pm3.27$    & $89.80$ \\ \hline
    FedOAP (Ours)                  & $\mathbf{94.39}$     & $\pm0.62$      & $\mathbf{95.31}$    & $\pm0.41$    & $\mathbf{87.22}$   & $\pm0.57$    & $\mathbf{92.31}$ \\ \hline
    \end{tabular}
    \end{center}
\end{table*}

\section{Experiments and Results}
    \subsection{Datasets} \label{sec:datasets_used}
    We evaluate FedOAP on three publicly available datasets encompassing diverse organs and imaging modalities. Specifically, we conduct tumor segmentation experiments on (i) dynamic contrast-enhanced magnetic resonance imaging (DCE-MRI) of the breast, (ii) structural MRI scans of the brain, and (iii) computed tomography (CT) images of the liver. To assess the cross-organ generalization capability of our framework, we further conduct evaluation on a dataset consisting of CT images of the lung, which is not included in the training phase.
    
    All input images and their corresponding ground truth segmentation masks are uniformly resized to a spatial resolution of $128 \times 128$. $10$\% of each dataset is reserved for testing, while $10$\% of the remaining data is allocated for validation. The rest is used for training the model. The detailed characteristics of each dataset are as follows:
    \begin{itemize}
        \item \textbf{BreastDM} dataset \citep{zhao2023breastdm} contains $232$ DCE-MRI scans of breast tissue, annotated for tumor presence. Among these, $147$ samples are malignant cases, while $85$ are benign. The DCE-MRI modality offers dynamic contrast information, which poses a distinct challenge for tumor boundary delineation due to its temporal nature. For our experiments, we ignore the benign cases and only work with the malignant ones.
        
        \item \textbf{BraTS 2020} dataset \citep{menze2014multimodal, bakas2017advancing, bakas2018identifying} consists of $369$ multi-contrast brain MRI scans annotated for glioma tumor sub-regions. Although the original BraTS dataset includes four imaging modalities: T1, T1-contrast (T1c), T2, and FLAIR, we aggregate the sequences and construct a single-channel input focused on whole tumor segmentation. This setup allows us to simplify modality dependency and emphasize anatomical variance across clients.

        \item \textbf{LiTS} dataset \citep{bilic2023liver} includes $201$ contrast-enhanced abdominal CT volumes with annotated liver tumors. For consistency with the other datasets and computational efficiency, each volume is preprocessed into two-dimensional slices and normalized as single-channel grayscale images.

        \item \textbf{Lung Cancer Segmentation dataset} \citep{nam2024lung} consists of $972$ lung CT slices and associated tumor segmentation masks. This dataset is used exclusively for out-of-distribution testing. It serves as a benchmark to evaluate the cross-organ generalizability of the proposed framework under domain and modality shifts.
    \end{itemize}

    \subsection{Evaluation Metrics}
    To evaluate the performance of our proposed FedOAP framework on tumor segmentation tasks, we use the mean of the Dice Similarity Coefficient (mDSC) \citep{milletari2016v} of the clients as the primary evaluation metric. Given a predicted mask $\hat{y} \in \{0, 1\}^{H \times W}$ and the corresponding ground truth ${y} \in \{0, 1\}^{H \times W}$, the Dice score is defined as:
    \begin{equation}
        \label{eq:diceScore}
        \operatorname{Dice}(\hat{y}, y) = \frac{2 \cdot \left|\hat{y} \cap y\right|}{\left|\hat{y}\right| + \left|{y}\right|}
    \end{equation}
    Here, $\left|\hat{y} \cap y\right|$ represents the correctly predicted tumor pixels, while $\left|\hat{y}\right|$ denote the total predicted tumor pixels and $\left|{y}\right|$ depict total true tumor pixels. Higher values of the dice score indicate a good overlap between predicted and true tumor regions.

    \subsection{Implementation Details} \label{sec:implementation}
    Our framework is implemented in Python $3.11.13$ with PyTorch $2.7.1$, CUDA $12.6$, and the Flower $1.4.0$ framework \citep{beutel2020flower} for client-server communication. Experiments are conducted on a single NVIDIA RTX A$4000$ GPU. We set the number of communication rounds to $\mathcal{T}=5$ with $\mathcal{K}=3$ clients, each training for one local epoch per round, followed by additional fine-tuning epochs $N_\epsilon=2$. The best-performing parameters are retained. Training uses AdamW \citep{loshchilov2017decoupled} with learning rate $\alpha=10^{-4}$, cosine annealing down to $10^{-6}$, and weight decay $10^{-5}$. Each minibatch has $16$ samples, and all random seeds are fixed to $42$ for reproducibility. For PBL, the threshold value is set to $\tau=0.75$, and the composite loss trade-off parameter $\lambda$ defaults to $0.1$. Justification about $\mathcal{K}$ included in Appendix \ref{sec:client_size}.

    \subsection{Comparison with State-of-the-Art Methods}
    \begin{figure*}[t]
        \centering
        \includegraphics[width=\textwidth]{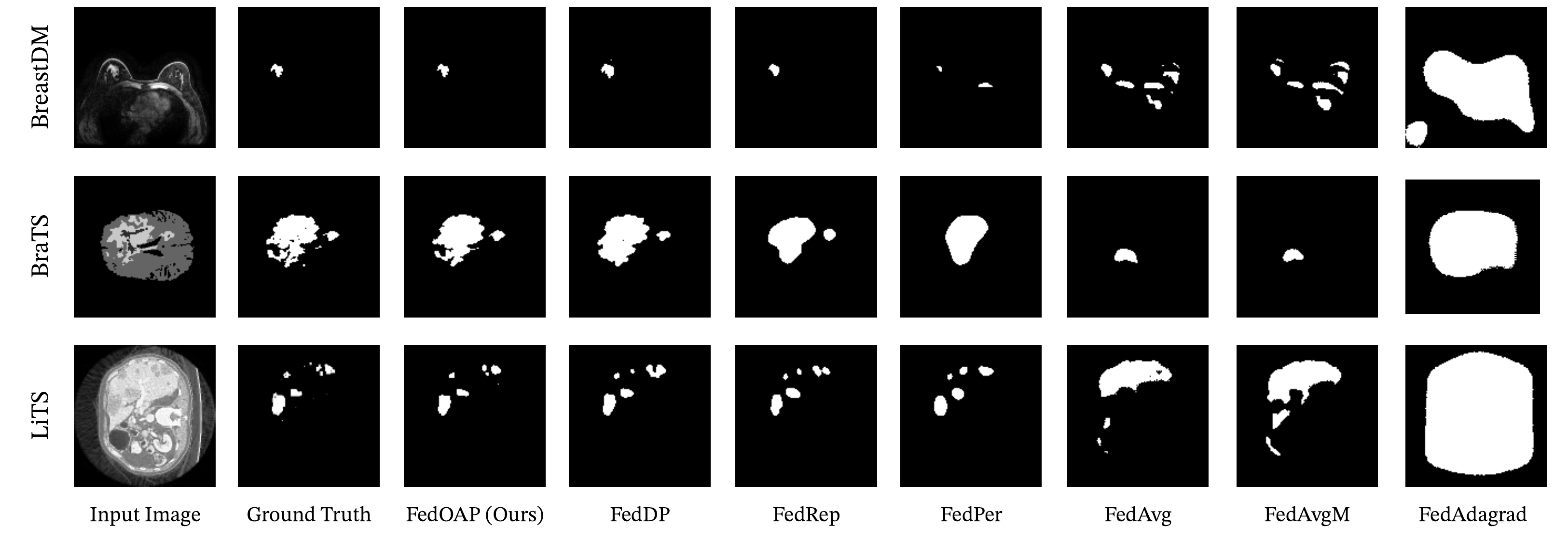} 
        \caption{Qualitative comparison of FedOAP with existing FL and PFL methods across different organ datasets. Each row depicts a random sample from a client, illustrating segmentation performance across methods.}
        \label{fig:compare}
    \end{figure*}
    We evaluate the performance of our proposed FedOAP framework against a suite of established federated learning (FL) and personalized federated learning (PFL) baselines. Among the FL baselines, we include the canonical FedAvg algorithm \citep{mcmahan2017communication} along with its notable extensions FedAvgM \citep{hsu2019measuring} and FedAdagrad \citep{reddi2020adaptive}. For PFL comparisons, we consider state-of-the-art methods such as FedPer \citep{arivazhagan2019federated}, FedRep \citep{collins2021exploiting}, and FedDP \citep{wang2023feddp}, which have demonstrated strong performance in heterogeneous data environments. To keep our comparisons fair, we evaluate each method under the same conditions and using the same hyperparameters.

    Table \ref{tab:results} presents the quantitative evaluation of all competing methods, demonstrating that the proposed FedOAP framework consistently outperforms both standard FL and PFL baselines across all datasets. It is evident that regular FL methods are not able to capture the cross-client and cross-domain feature space, resulting in their poor performance. We can see FedAvg typically outperforming FedAvgM and FedAdagrad. These results highlight the inherent limitations of global model aggregation in the presence of strong data heterogeneity. Notably, while most FL methods perform poorly on the BraTS and LiTS datasets, FedAdagrad exhibits relatively higher accuracy on BraTS. Looking at most of the results, FedAvg and FedAvgM mostly fail to identify even a single tumor pixel whereas FedAdagrad barely identifies a basic outlined shape. This could be attributed to the presence of larger but structurally complex tumors, where simple global updates may suffice to capture coarse tumor characteristics without requiring precise boundary delineation.
    
    The PFL methods are comparatively better as they are specifically tailored to personalize every client to its data distribution. FedPer and FedRep both achieve significant gains when compared to FL methods. But their high inter-client variance points to an inability to handle cross-client knowledge transfer. Both FedDP and FedOAP (ours) exhibit superior results across different clients. This further supports the ability of the attention mechanism to learn from a shared representation. Furthermore, our method outperforms FedDP because of our focus in perfecting the prediction boundaries using noisy supervision. The benefits of this is particularly evident in the BraTS dataset, where complex tumor morphologies demand fine-grained spatial awareness. Additionally, our proposed personalized cross-attention mechanism enables clients to integrate context-aware features from different organ domains while preserving privacy through local query retention.

    To qualitatively analyze the performance, we visualize the segmentation results of the methods in Figure \ref{fig:compare}. For a randomly selected instance from each dataset, we display the input image, corresponding ground truth mask, and the predicted segmentation masks generated by the competing methods. Nearly all the methods have a difficult time delineating complex tumor boundaries, particularly in anatomically diverse scenarios. In contrast, the proposed FedOAP framework demonstrates a notable ability to adapt to intricate spatial patterns and generate precise segmentation masks. The comparatively tiny tumor regions in the BreastDM dataset is difficult to capture for the FL methods. Only the personalized models have localized these regions with varying levels of accuracy. This further supports the claim of how high data heterogeneity can devastate model performance. Although they may have localized the regions, most PFL models struggle with predicting the irregular boundaries. In comparison, FedOAP achieves superior boundary adherence and spatial precision. This is particularly evident in samples from both the BraTS and LiTS datasets, where tumors display substantial morphological variability. These results affirm the efficacy of our boundary-aware optimization strategy and decoupled cross-attention mechanism, which combined enable robust and context-sensitive segmentation.
    \begin{table}[t]
        \caption{Generalization results comparing FedOAP with FedDP on the Lung Cancer Segmentation dataset. Best scores are in bold.}
        \label{tab:zeroShot}
        \begin{center}
        \begin{tabular}{c|cc}
        \hline
        \multirow{2}{*}{Method} & \multicolumn{2}{c}{ Dice(\%) $\uparrow$} \\ \cline{2-3} 
                                & Zero-Shot         & Fine-Tuned         \\ \hline
        FedDP                   & $1.1$                   & $65.4$             \\
        FedOAP (Ours)                  & $\mathbf{7.1}$                   & $\mathbf{72.3}$             \\ \hline
        \end{tabular}
        \end{center}
    \end{table}
    \subsection{Generalization on Unseen Data}
    As FedOAP aims to leverage a shared representation across clients and incorporate organ-specific features from to enhance its own predictions, we hypothesize that it should perform significantly better when tested on a data distribution that was unseen during training. To empirically validate our hypothesis, we compare FedOAP (ours) against FedDP, the best PFL framework in previous evaluations, on a completely unseen lung cancer segmentation dataset \citep{nam2024lung}. The results of our experiment are reported in Table \ref{tab:zeroShot}. Initially, we evaluate both FedOAP and FedDP on the new dataset in a zero-shot setting. And both of them achieve a low dice score ($1.1$\% for FedDP and $7.1$\% for FedOAP) as they are not personalized to the new domain. However, FedOAP consistently outperforms FedDP, demonstrating its capacity to exploit shared representations via the proposed DCA mechanism, even without client-specific tuning. Finally, we fine-tune both methods on the new dataset for only two epochs without any inter-client parameter exchange. This simulates a realistic scenario where a new client adapts its model using only previously shared global knowledge. Post fine-tuning, FedOAP achieves a noticeable performance improvement and outperforms FedDP by a significant margin, thereby affirming the effectiveness of its design in enabling rapid adaptation through robust feature generalization.

    \subsection{Ablation Study}
    We thoroughly investigate the individual and cumulative contribution of each proposed component, namely DCA, spatial adapter, and PBL. To this end, we incrementally incorporate these modules into the base framework and report the mean Dice Similarity Coefficient (mDSC\%) across all clients in Table \ref{tab:ablation}. In the baseline setting, none of the proposed enhancements are included. Even in this configuration, our framework achieves competitive performance on the BreastDM dataset. This observation may be attributed to the well-defined tumor boundaries in DCE-MRI breast images, which makes the segmentation task less reliant on complex contextual modeling.

    \begin{table*}[t]
        \caption{Ablation study showing mean Dice scores (\%)$\uparrow$ for the proposed components: DCA, spatial adapter, and PBL across datasets. Best results are highlighted in bold.}
        \label{tab:ablation}
        \begin{center}
        \setlength{\tabcolsep}{10pt}
        \begin{tabular}{@{}cccccc@{}}
        \hline
        DCA      & Spatial Adapter & PBL      & BreastDM & BraTS  & LiTS   \\ \hline
        $\times$ & $\times$        & $\times$ & $85.1$   & $25.9$ & $5.3$  \\
        \checkmark         & $\times$        & $\times$ & $88.9$   & $52.7$ & $8.3$  \\
        \checkmark         & \checkmark                & $\times$ & $89.99$  & $65.8$ & $86.1$ \\
        \checkmark         & \checkmark                & \checkmark         & $\mathbf{94.4}$   & $\mathbf{95.3}$ & $\mathbf{87.3}$ \\ \hline
        \end{tabular}
        \end{center}
    \end{table*}
    The incorporation of the proposed DCA mechanism yields a marked improvement in performance on the BraTS dataset. This empirical finding reinforces our hypothesis that leveraging cross-organ feature representations substantially enhances the model’s ability to capture complex structural patterns, particularly in cases involving intricate brain tumor morphology. In contrast, the performance gains observed on the BreastDM and LiTS datasets are relatively modest, suggesting that the benefits of DCA are most pronounced where the target structures exhibit high variability and spatial complexity.

    The inclusion of the spatial adapter significantly improves the performance of the challenging LiTS dataset (from $8.6$\% to $86.1$\%). This underscores the critical importance of client-specific adaptation in effectively handling the low-contrast, structurally cluttered, and highly heterogeneous nature of abdominal CT scans. Additionally, all the other clients possess data from a different modality (MRI) when compared to the LiTS dataset, which contains CT images, further adding to its heterogeneity. A moderate performance gain is also observed on the BraTS dataset, suggesting that personalized modeling facilitates improved localization in structurally variable brain tumor regions. In contrast, the performance on the BreastDM dataset remains relatively unchanged, indicating a potential performance plateau wherein further personalization yields diminishing returns.

    Finally, we include PBL to complete the set of proposed enhancements. This yields a marked improvement in segmentation performance on the BraTS dataset, attributed to the noisy supervision mechanism which encourages the model to better delineate complex tumor boundaries seen on brain MRI scans. Similarly, the BreastDM dataset benefits from enhanced boundary precision, resulting in more accurate segmentation masks for the comparatively small and well-defined tumor regions. However, the performance gain on the LiTS dataset remains marginal, attributed to its heavy reliance on client-specific tuning.

    \subsection{Efficiency Analysis}
    Table \ref{tab:transmission} depicts the transmission cost between the clients and the server for each communication round.
    \begin{table}[t]
        \caption{Comparison of transmission overhead (in MB) per communication round between a client and server for different federated learning methods.}
        \label{tab:transmission}
        \begin{center}
        \begin{tabular}{cc}
            \hline
            Method        & Transmission Size (MB) \\ \hline
            FedAvg        & $\sim 118.37$          \\
            FedAvgM       & $\sim 118.37$          \\
            FedAdagrad    & $\sim 118.37$          \\ \hline
            FedPer        & $\sim 71.88$           \\
            FedRep        & $\sim 71.88$           \\
            FedDP         & $\sim 130.44$          \\ \hline
            FedOAP (Ours) & $\sim 130.44$          \\ \hline
        \end{tabular}
        \end{center}
    \end{table}
    FedOAP introduces a modest increase in transmission overhead relative to FedPer and FedRep while delivering superior performance. The lower transmission cost of these methods arises from their design choice of transmitting only the encoder parameters to the server while retaining the decoder locally on each client. In contrast, FedOAP personalizes only the query vectors $\mathbf{q_k}$ and the spatial adapter ${\phi}_k$, with all remaining parameters shared with the server. When compared to FedAvg, FedAvgM, and FedAdagrad, FedOAP incurs slightly higher transmission cost, as it requires each client to communicate the concatenated key–value representations aggregated from all participants, whereas the baselines only transmit parameters from a single client. The key observation is that FedOAP achieves transmission overheads on par with FedDP while consistently outperforming it in model performance. Additional experiments about efficiency analysis are provided in section \ref{sec:efficiency_analysis} of the supplementary document.

\section{Conclusion}
We introduced FedOAP, a PFL framework for organ-agnostic tumor segmentation that combines decoupled cross-attention with a boundary-aware loss. FedOAP consistently outperforms existing FL and PFL methods in accuracy and generalization across heterogeneous and unseen domains. Future work will explore its scalability with larger datasets and under irregular federated conditions.


\bibliography{ref}

\clearpage
\appendix
\thispagestyle{empty}

\onecolumn

\section{Related Work}\label{sec:related_work}
Federated learning (FL) has garnered significant attention as a decentralized paradigm for collaboratively training models across distributed clients without sharing raw data \citep{horst2025federated, mcmahan2017communication, konevcny2016federated, li2020federated}. The foundational algorithm, FedAvg (Federated Averaging) proposed by \citep{mcmahan2017communication}, serves as a baseline approach for federated training. However, recent works such as \citep{karimireddy2020scaffold, collins2021exploiting, li2021model} show how the model convergence is effected in heterogeneous settings caused by local updates on non-IID data. FedAvg has been modified in \citep{hsu2019measuring, reddi2020adaptive, li2020federated} to address the client drift issue, focused on learning a singular global model while also being generalizable enough to suit the needs of heterogeneous clients. Nonetheless, relying solely on a global model often fails to adequately capture the diverse and client-specific patterns inherent in highly non-IID data distributions \citep{zhao2018federated}.

Personalized Federated Learning (PFL) addresses the limitations of global models by learning client-specific models tailored to local data distributions \citep{tan2022towards}. Approaches such as \citep{arivazhagan2019federated} introduce client-specific layers, while others like \citep{collins2021exploiting} focus on learning a shared representation across heterogeneous clients. Parameter decoupling strategies have also been explored, including FedBN \citep{li2021fedbn}, which personalizes batch normalization layers, and FedRep or FedBABU \citep{collins2021exploiting, oh2021fedbabu}, which personalize prediction layers. Despite these advancements, the use of cross-attention in PFL remains largely underexplored. Given its effectiveness in modeling long-range dependencies and complex feature interactions \citep{wang2023feddp}, attention based methods present a promising direction for enhancing both global representation sharing and local adaptation. While recent PFL approaches have shown progress in medical image segmentation \citep{chang2020synthetic, dong2021federated, kaissis2020secure, jiang2023iop, wang2022personalizing, liu2021feddg, dai2024federated, wang2023feddp}, few consider the benefits of leveraging cross-organ feature representations to improve local predictions.

Inconsistencies between predicted masks and ground truth offer a valuable signal for enhancing model supervision, especially for fine-grained boundary refinement. These subtle yet critical differences can impair segmentation performance, yet few existing methods leverage them for model calibration. \citep{wang2022personalizing} propose a local calibration framework that incorporates inconsistency-aware supervision at both the feature and prediction levels. Their method combines contrastive site embeddings with channel selection and attention-based refinement to better capture anatomically ambiguous boundaries, albeit at increased inference cost. Similarly, \citep{wang2023feddp} introduce an inconsistency-guided calibration strategy that computes inter-site prediction disagreements to inform additional supervision. While effective, this method incurs significant memory overhead by requiring access to multiple clients' predictions or model weights.

\section{Additional Efficiency Analysis} \label{sec:efficiency_analysis}
    \subsection{Computational Overhead}
    Table \ref{tab:flops} compares the computational complexity of FedOAP with FedDP. The total FLOPs capture the overall training cost aggregated across all clients, while the Attention Mechanism column isolates the computational load attributed to Local Query (LQ) introduced in \citet{wang2023feddp} and the proposed Decoupled Cross-Attention (DCA). Both mechanisms exhibit comparable FLOP requirements and scale linearly with the number of clients, indicating similar efficiency. Despite this equivalence, FedOAP achieves superior performance due to the incorporation of cross-attention, enabling richer feature exchange across distributions.
    \begin{table}[h]
        \caption{Comparison of FLOP counts between FedOAP and FedDP across clients for the training stage.}
        \label{tab:flops}
        \begin{center}
        \begin{tabular}{c|cc}
            \hline
            \multirow{2}{*}{Method} & \multicolumn{2}{c}{GFLOPs}    \\ \cline{2-3} 
                                    & Attention Mechanism & Total   \\ \hline
            FedDP                   & $\sim8.73$              & $\sim197.5$ \\
            FedOAP (Ours)           & $\sim8.73$              & $\sim232.6$ \\ \hline
        \end{tabular}
        \end{center}
    \end{table}
    
    The slightly higher total FLOP count observed for FedOAP arises from the inclusion of the spatial adapter ${\phi}_k$, which is not present in FedDP. This adapter facilitates personalized adaptation to local data distributions while remaining client-specific, thereby mitigating client drift during parameter aggregation \citep{arivazhagan2019federated, collins2021exploiting}. Overall, while FedOAP introduces a slight increase in total computational cost, this overhead is justified by its improved generalization and personalization efficiency across heterogeneous clients.
    
    Table \ref{tab:memory} presents a comparison of the memory requirements for boundary-guided edge refinement mechanisms within a single client. \citet{wang2023feddp} proposed the Inconsistency Guided Calibration (IGC) module, which enhances predictions by comparing the outputs of multiple client models on identical inputs to identify inconsistency regions. These inconsistencies are then used to refine each client’s prediction. In contrast, our proposed Perturbed Boundary Loss (PBL) introduces additive noise to the inconsistent regions between the prediction and the corresponding ground truth, thereby improving boundary precision without relying on inter-client information exchange.
    \begin{table}[h]
        \caption{Comparison of memory overhead (in MB) for boundary-guided refinement mechanisms per client.}
        \label{tab:memory}
        \begin{center}
        \begin{tabular}{cc}
            \hline
            Method     & Memory Required (MB) \\ \hline
            IGC        & $\sim4.23$           \\
            PBL (Ours) & $\sim3.5$            \\ \hline
        \end{tabular}
        \end{center}
    \end{table}

    IGC exhibits substantially higher memory consumption compared to PBL. This increase arises because IGC requires loading and running inference using all client-specific models simultaneously on a single client to compute inter-client inconsistencies. Conversely, PBL only compares the local model’s output with its ground truth labels, which significantly reduces memory usage. Moreover, IGC is less effective in cross-organ scenarios where client data originate from distinct organ distributions, leading to excessively large inconsistency regions when comparing to predictions from other clients. By grounding the comparison within each client’s local domain, PBL achieves more stable boundary refinement while maintaining lower memory overhead.

    \subsection{Analysis of Training and Inference Overheads}
    \begin{table}[h]
        \caption{Comparison of FLOP counts between FedOAP and FedDP across clients for the training stage.}
        \label{tab:time}
        \begin{center}
        \begin{tabular}{c|ccc}
            \hline
            \multirow{2}{*}{Method} & \multicolumn{3}{c}{Time Taken (s)}                 \\ \cline{2-4} 
                                    & Federated Training & Local Fine-tuning & Inference \\ \hline
            FedAvg                  & $400.45$           & -                 & $6.11$    \\
            FedAvgM                 & $401.32$           & -                 & $6.21$    \\
            FedAdagrad              & $393.42$           & -                 & $6.29$    \\
            FedPer                  & $421.31$           & -                 & $6.77$    \\
            FedRep                  & $622.95$           & $115.41$          & $7.19$    \\
            FedDP                   & $496.71$           & $248.27$          & $7.49$    \\ \hline
            FedOAP (Ours)                  & $621.77$           & $176.5$           & $8.16$    \\ \hline
        \end{tabular}
        \end{center}
    \end{table}

\begin{figure}[ht]
        \centering
        \includegraphics[width=0.6\columnwidth]{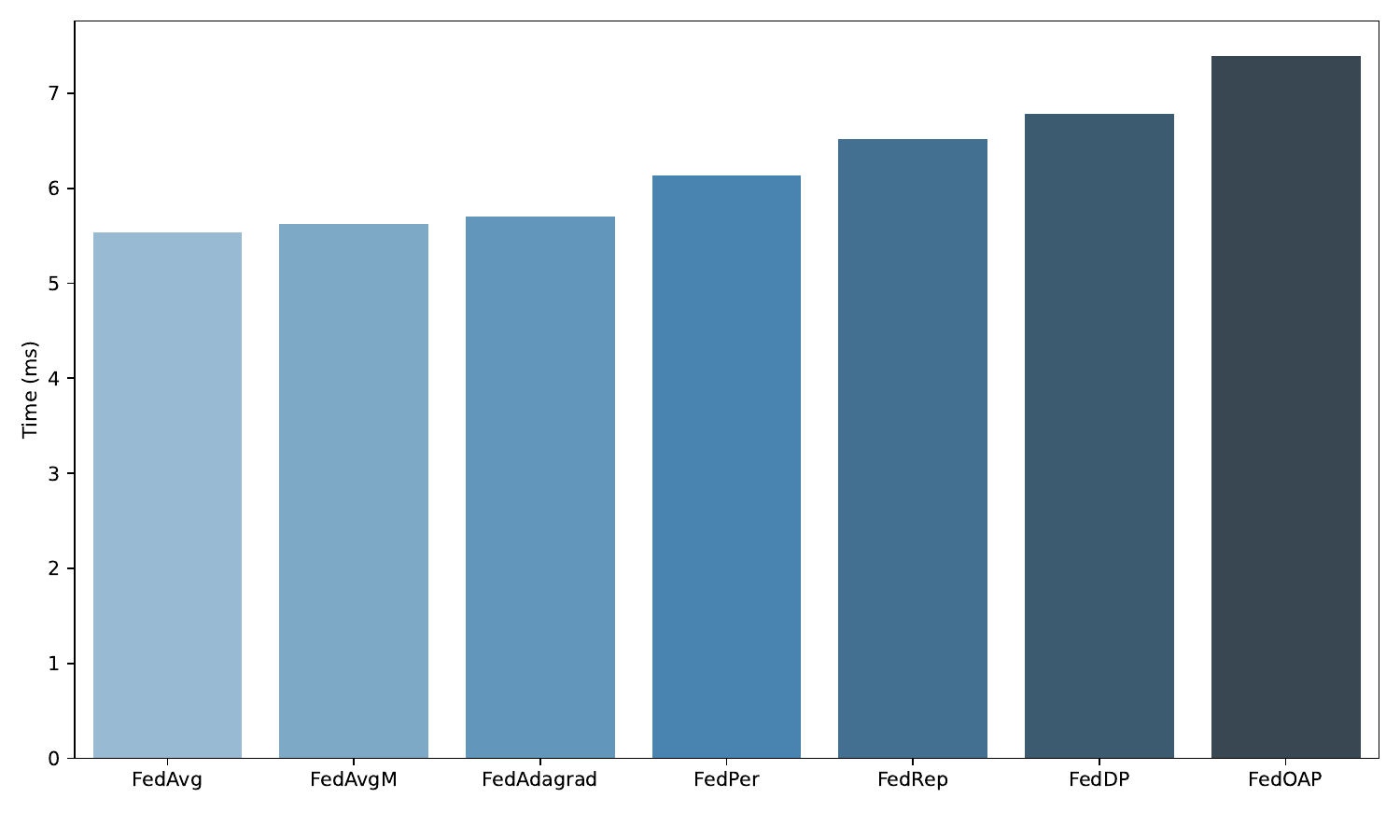} 
        \caption{Average inference time (in ms) per sample for each FL method.}
        \label{fig:inference_time}
    \end{figure}

    Table \ref{tab:time} presents the total training, local fine-tuning and inference time for different federated learning methods for the experimental setting described in Section \ref{sec:implementation}. FedOAP incurs higher training and inference times compared to other other learning methods. This could be attributed to having higher FLOP count as seen previously. The architecture of our method slightly increases the number of parameters \textit{i.e.} having local spatial adapters. This can slightly affect the performance of our method. This trend is also noticeable when comparing the inference time per sample (tested with $1104$ samples) as shown in Figure \ref{fig:inference_time}. However, FedOAP took considerably less time in the local fine-tuning step, ensuring efficient transfer of knowledge to a specific data distribution from a shared feature space.

\section{Experimental Parameters} \label{sec:client_size}
    \subsection{Number of Clients}
    In our experimental setup, we set the number of clients to $\mathcal{K}=3$, which differs from conventional federated learning (FL) configurations that typically involve a larger number of clients to simulate real-world network heterogeneity. However, in our study, each client represents an independent clinical institution or repository, where data is sourced from a distinct organ domain. Specifically, we utilize three datasets: BreastDM \citep{zhao2023breastdm}, BraTS \citep{menze2014multimodal, bakas2017advancing, bakas2018identifying}, and LiTS \citep{bilic2023liver}; corresponding to breast, brain, and liver tumor segmentation tasks, respectively. Assigning one dataset per client allows us to model an organ-distributed federated scenario, where data heterogeneity naturally arises from domain differences rather than patient-level variations. This design choice enables a controlled yet realistic evaluation of the proposed FedOAP framework’s ability to generalize across organ-specific feature distributions, while maintaining the interpretability of cross-organ knowledge aggregation.

    \subsection{Communication Rounds and Local Training}
    \begin{figure}[ht]
        \centering
        \includegraphics[width=0.6\columnwidth]{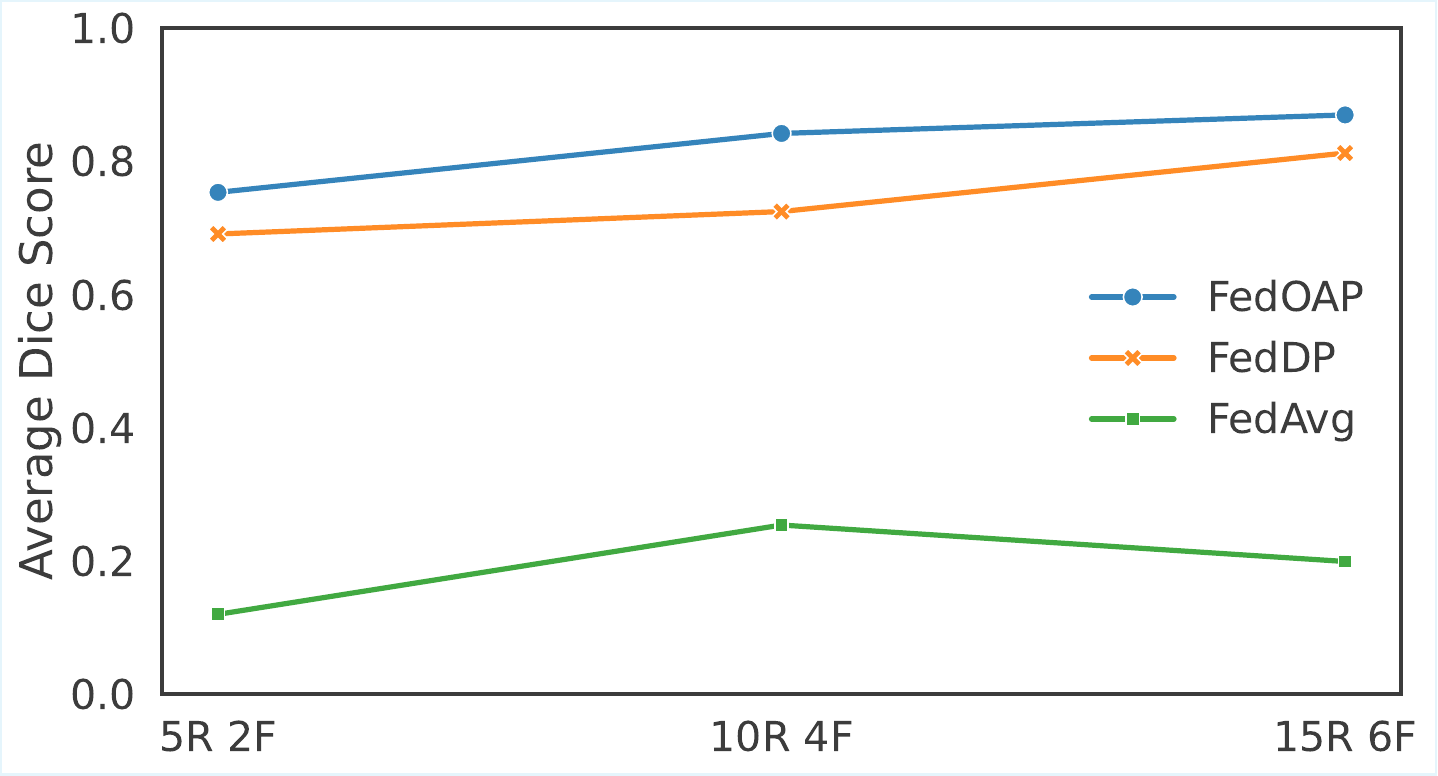} 
        \caption{Comparison of FedOAP using $\mathcal{K}=30$ in various training scenarios. Here, $R$ denotes the number of communication rounds and $F$ denotes the number of local fine-tuning epochs.}
        \label{fig:more_clients}
    \end{figure}

    In our experiments, the total number of federated communication rounds is set to $\mathcal{T}=5$, followed by $N_{\epsilon}=2$ local fine-tuning epochs. This configuration was chosen based on the observed convergence behavior of FedOAP. Empirically, FedOAP demonstrates substantially faster convergence compared to other federated and personalized federated learning baselines, achieving stable performance within only a few global rounds. The decoupled cross-attention mechanism in FedOAP facilitates more effective knowledge aggregation across clients, enabling the model to reach optimal performance without extensive communication. Furthermore, limiting the number of global rounds reduces the overall computational and communication overhead, which is desirable in realistic federated settings. The additional fine-tuning epochs allow each client to refine its personalized parameters on local data quickly, ensuring improved adaptation to organ-specific characteristics without incurring significant training cost.

    To further assess the scalability and robustness of FedOAP, we extend our experiments to a larger federated setting with $\mathcal{K}=30$ clients. The BreastDM, BraTS, and LiTS datasets are evenly partitioned into multiple subsets to accommodate the increased number of clients while maintaining balanced data distribution. We evaluate three distinct training paradigms, defined by $(\mathcal{T}=5, N_{\epsilon}=2)$, $(\mathcal{T}=10, N_{\epsilon}=4)$, and $(\mathcal{T}=15, N_{\epsilon}=6)$. The results, illustrated in Figure \ref{fig:more_clients}, clearly demonstrate that FedOAP continues to outperform all baseline methods across different configurations. Notably, FedOAP achieves faster convergence with fewer communication rounds, highlighting the effectiveness of its shared feature space and spatial adapter modules in mitigating client drift and improving generalization. In contrast, methods such as FedAvg exhibit increasing divergence with prolonged training, underscoring the stability and efficiency advantages of our proposed framework.

\section{Justification for Organ-Agnostic Claim}
The term organ-agnostic in FedOAP reflects the framework’s ability to generalize and adapt across diverse organ domains relying on minimal organ-specific training. This characteristic is seen in Table \ref{tab:zeroShot} through the model’s superior zero-shot performance, achieving a $6$\% improvement over FedDP under cross-organ evaluation. Such improvement demonstrates FedOAP’s enhanced transferability and capacity to learn a shared, domain-invariant feature representation that generalizes effectively to unseen organ distributions. Furthermore, when fine-tuned locally for only two additional epochs, FedOAP exhibits a substantial performance boost, surpassing all competing methods by a notable margin. This rapid adaptation underscores the strength of its transferable representations and the efficacy of its decoupled cross-attention and spatial adaptation mechanisms. Hence, the framework’s consistent cross-domain generalization and efficient personalization together justify its designation as an organ-agnostic federated learning paradigm.

\section{Model} \label{sec:model}
To implement our proposed method and test all other FL and PFL methods, we employ a modified U-Net backbone \citep{ronneberger2015u} augmented with two key components: a Cross-Attention module and a Spatial Adapter. This architecture is designed to facilitate cross-organ feature interaction and domain-adaptive refinement while maintaining computational efficiency. The base architecture follows the conventional U-Net encoder–decoder structure, comprising symmetric downsampling and upsampling paths with skip connections to preserve spatial information.

The encoder begins with an initial double convolution block comprising two sequential $3 \times 3$ convolutional layers, each followed by instance normalization and ReLU activation. Feature channels are progressively expanded through three Down blocks using $2 \times 2$ max pooling and subsequent double convolutions, with channel dimensions increasing from $64$ to $512$. The bottleneck consists of two parallel $3 \times 3$ convolutional stages with $1024$ channels: one serving as the query branch and the other as the key–value branch for cross-attention. The cross-attention module performs spatial feature alignment between the query and key–value representations using $1 \times 1$ convolutions to generate queries, keys, and values, followed by multi-head attention and normalization via group normalization. This module also allows cross-client feature exchange, enhancing generalization across organ domains.

The decoder mirrors the encoder through four Up blocks, each using a $2 \times 2$ transposed convolution for upsampling, concatenation with the corresponding encoder feature map, and a subsequent double convolution to refine spatial details. Channel dimensions are reduced progressively from $1024$ back to $64$. At the final stage, a spatial adapter module composed of two $3 \times 3$ convolutional layers with residual connections—further refines spatial consistency and preserves localized texture information. A final $1 \times 1$ convolution projects the $64$-channel feature map to the desired segmentation mask.

\section{Dataset Preparation} \label{sec:datasets}
We have already provided  a description of the datasets used in our experimentation in Section \ref{sec:datasets_used}. As we are dealing with data from multiple input modalities, we unify them by converting all ground truth to a single channel $2$ dimensional input. $20$\% of the data from each dataset is used for testing. $20$\% of the remaining data is used for validation during training while the rest is kept for training.

\section{Segmentation Loss}
We already discussed the perturbed loss $\mathcal{L}_p$ in Section \ref{sec:loss}. Our composite objective function additionally depends on a segmentation loss $\mathcal{L}_s$. To ensure accurate delineation of organ boundaries during segmentation, we employ a loss function that combines the strengths of Binary Cross-Entropy (BCE) and the Dice coefficient. The BCE component penalizes pixel-wise misclassifications, promoting accurate foreground–background discrimination, while the Dice component directly optimizes the overlap between predicted and ground truth masks, enhancing spatial consistency in regions of fine structure.

Formally, given a predicted probability map $\hat{y}_{i,k}$ and its corresponding ground truth mask $y_{i,k}$, the segmentation loss $\mathcal{L}_s$ is defined as a weighted combination of the two terms:
\begin{equation}
    \label{eq:segLoss}
    \mathcal{L}_s = 0.5 \times \mathcal{L}_{BCE}(\hat{y}_{i,k},y_{i,k}) + \mathcal{L}_{Dice}(\hat{y}_{i,k},y_{i,k})
\end{equation}
where the BCE loss operates on the logits, and the Dice term measures the degree of spatial overlap between prediction and ground truth. This joint formulation enables stable optimization across varying organ sizes and image resolutions while maintaining sensitivity to small structures.

\end{document}